\documentclass[10pt,twocolumn,letterpaper]{article}

\usepackage{3dv}
\usepackage{times}
\usepackage{epsfig}
\usepackage{graphicx}
\usepackage{amsmath}
\usepackage{amssymb}
\usepackage{multirow}
\usepackage{booktabs}
\usepackage{authblk}
\usepackage{enumitem}

\usepackage[pagebackref=true,breaklinks=true,letterpaper=true,colorlinks,bookmarks=false]{hyperref}

\threedvfinalcopy % *** Uncomment this line for the final submission

\def\threedvPaperID{377} % *** Enter the 3DV Paper ID here

\ifthreedvfinal\fi
\begin{document}

% %%%%%%%%% TITLE
% \title{Correspondence Matrices are Underrated}
% % it is a double blind review. Dont add author names
% \author{Tejas Zodage\\
% Carnegie Mellon University\\
% Pittsburgh, PA, USA\\
% {\tt\small tzodage@andrew.cmu.edu}
% % % For a paper whose authors are all at the same institution,
% % % omit the following lines up until the closing ``}''.
% % % Additional authors and addresses can be added with ``\and'',
% % % just like the second author.
% % % To save space, use either the email address or home page, not both
% \and
% Rahul Chakwate\\
% Indian Institute of Technology Madras\\
% Chennai, TN, India\\
% % {\tt\small ae16b005@smail.iitm.ac.in}
% \and
% Vinit Sarode\\
% Carnegie Mellon University\\
% Pittsburgh, PA, USA\\
% % {\tt\small vsarode@andrew.cmu.edu}
% \and
% Arun Srivatsan \\
% Carnegie Mellon University\\
% Pittsburgh, PA, USA\\
% % {\tt\small rarunsrivatsan@gmail.com }
% \and
% Howie Choset\\
% Carnegie Mellon University\\
% Pittsburgh, PA, USA\\
% % {\tt\small choset@andrew.cmu.edu}
% }
% \maketitle
% %\thispagestyle{empty}

%%%%%%%%% TITLE
% \title{Point Cloud Registration: Correspondence Matrices are Underrated}
\title{Correspondence Matrices are Underrated}
\author[1]{Tejas Zodage *}
% it is a double blind review. Dont add author names

% Carnegie Mellon University\\
% Pittsburgh, PA, USA\\
% {\tt\small tzodage@andrew.cmu.edu}
% \and
% Rahul Chakwate\\
% Indian Institute of Technology Madras\\
% Chennai, TN, India\\
% % {\tt\small ae16b005@smail.iitm.ac.in}
% \and
% Vinit Sarode\\
% Carnegie Mellon University\\
% Pittsburgh, PA, USA\\
% % {\tt\small vsarode@andrew.cmu.edu}
% \and
% Arun Srivatsan \\
% Carnegie Mellon University\\
% Pittsburgh, PA, USA\\
% % {\tt\small rarunsrivatsan@gmail.com }
% \and
% Howie Choset\\
% Carnegie Mellon University\\
% Pittsburgh, PA, USA\\
% {\tt\small choset@andrew.cmu.edu}
% }
\author[2]{Rahul Chakwate}
\author[1]{Vinit Sarode}
\author[1]{Rangaprasad Arun Srivatsan}
\author[1]{Howie Choset}
\affil[1]{Carnegie Mellon University }
\affil[2]{Indian Institute of Technology, Madras }
\affil[*] {\tt tzodage@andrew.cmu.edu}
\maketitle
%\thispagestyle{empty}

%%%%%%%%% ABSTRACT

\begin{abstract}
%DO NOT TOUCH:
Point-cloud registration (PCR) is an important task in various applications such as robotic manipulation, augmented and virtual reality, SLAM, etc. PCR is an optimization problem involving minimization over two different types of interdependent variables: transformation parameters and point-to-point correspondences. Recent developments in deep-learning have produced computationally fast approaches for PCR. The loss functions that are optimized in these networks are based on the error in the transformation parameters. We hypothesize that these methods would perform significantly better if they calculated their loss function using correspondence error instead of only using error in transformation parameters. We define correspondence error as a metric based on incorrectly matched point pairs. We provide a fundamental explanation for why this is the case and test our hypothesis by modifying existing methods to use correspondence-based loss instead of transformation-based loss. These experiments show that the modified networks converge faster and register more accurately even at larger misalignment when compared to the original networks.

\end{abstract}

%%%%%%%%% BODY TEXT
\section{Introduction}
{\let\thefootnote\relax\footnote{ We would like to thank Center of Machine Learning and Health, CMU for providing the \href{https://www.cs.cmu.edu/cmlh-2020-fellows}{CMLH fellowship, 2020} to Mr. Tejas Zodage and thus partially supporting this work.}}
{\let\thefootnote\relax\footnote{ Code available at: \href{https://github.com/tzodge/PCR-CMU}{\tt https://github.com/tzodge/PCR-CMU} }}
% Developments of sensors such as LIDAR, RealSense, Kinect etc. have given computers the ability to analyze the scenes in 3-dimensions as opposed to conventional 2-dimensional images. The sensors output the scene information in the form of sets of 3D unordered points collectively known as point clouds. Due to limited field of view and/or physical constraints such as occlusions, the point clouds of a scene are collected in patches from independent frames. These individual patches need to be aligned in order to gain complete scene-understanding. 

%  augmented reality ~\cite{AR_icp}, scene understanding for self driving cars~\cite{chen20203d}, simultaneous localisation and mapping~(SLAM)~\cite{holz2010sancta}

% The task of aligning point clouds obtained from different frames is known as point cloud registration~(PCR). PCR is a crucial step in applications such as 3D reconstruction~\cite{newcombe2011kinectfusion}, robot manipulation~\cite{xiang2017posecnn}, etc. Conventionally PCR is posed as an optimization problem with two interdependent parameters: transformation and correspondence. Various optimization techniques have been proposed in order to make this optimization process fast and accurate. However, recently developed deep-learning-based PCR methods such as PCRNet~\cite{PCRNet}, DCP ~\cite{DCP}, PointNetLK ~\cite{PointNetLK} have shown faster and more accurate results than the conventional PCR methods.

Point cloud registration~(PCR), the task of finding the alignment between pairs of point clouds is often encountered in several computer vision~\cite{newcombe2011kinectfusion, PointNetLK} and robotic applications~\cite{aldoma2012our,Xiang-RSS-18,rusu2009detecting,segal2009generalized}. 
% Popular registration algorithms such as iterative closest point (ICP)~\cite{ICP} and its numerous variants~\cite{rusinkiewicz2001efficient} approach this task by iteratively finding pose parameters and point correspondences
A critical aspect of registration is determining a correspondance \emph{i.e.} mapping between points of first cloud to the points of the second. Most approaches to registration (e.g., ~\cite{ICP}) use simple rules-of-thumb or implement a separate procedure to establish correspondances.
While these approaches have been widely used, they do suffer from computational complexity impacting their performance to determine pose parameters in real-time. Recent developments in deep learning-based registration approaches have resulted in faster and, under some circumstances, more accurate results~\cite{PointNetLK,PCRNet,DCP, gojcic20193DSmoothNet}.

\begin{figure}[t]
    \begin{center}
        \includegraphics[width=0.7\linewidth]{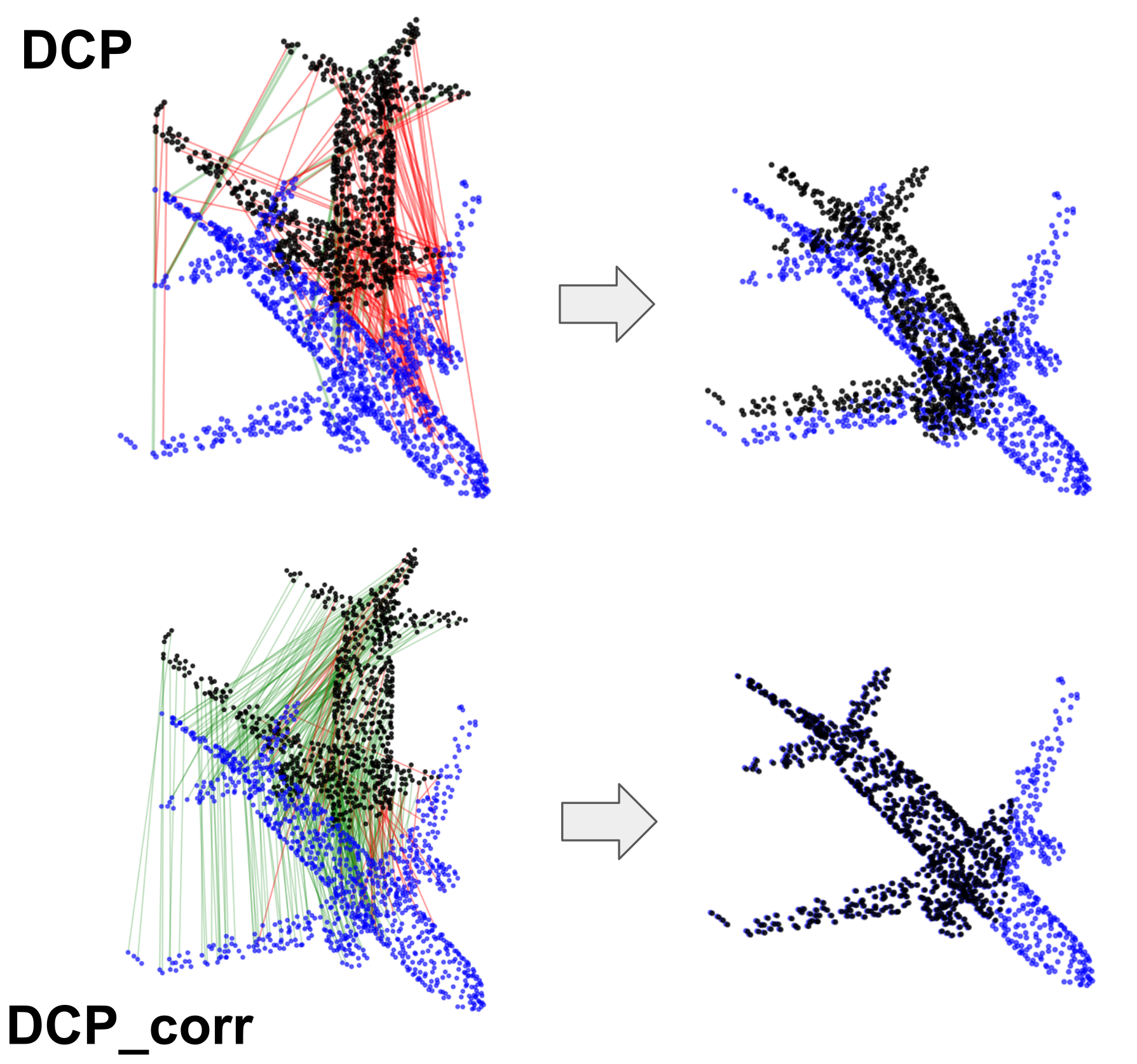}  
    \end{center}
    \caption{When learning to predict the pose parameters, deep-learning-based methods such as DCP~\cite{DCP} learn correspondence (mapping between the points of point clouds) implicitly. If the same networks are trained to explicitly learn correspondence (DCP\_corr), the resulting registration is more accurate. Template point cloud is shown in blue, source point cloud in black. Green arrows show correct correspondence. Red arrows show incorrect correspondence.} 
    % \label{fig:long}
    \label{fig:intro_result}
\end{figure}

 Unlike conventional approaches, most deep learning approaches directly estimate the pose and often do not explicitly estimate point correspondences. Instead, they implicitly learn the correspondences while being trained. While exploring the relation between correspondence and registration, we observed that perturbing the correspondence produced only small changes in the final pose estimation when compared to perturbations in the axis-angle representation of the rotation~(see section \ref{robustness_correspondence} for more details). % In other words, the correspondence parameters were more numerically stable than the final pose parameters and therefore underrrated as a basis for learning to register point clouds. 

% Deep-learning-based methods generally predict transformation between input point clouds. Usually the training procedure involves loss function defined based on predicted transformation and the ground-truth transformation. Such networks have shown a registration accuracy of $\approx 3^\circ$ on ModelNet40 dataset for pointclouds misaligned in the range $[-45^\circ,+45^\circ]$. Also these networks are not able to register partial point clouds(Need better wording here.) 
Based on this observation, we hypothesize that higher registration accuracy can be achieved by training the networks to explicitly predict point correspondences instead of implicitly learning them.
In order to test this hypothesis, we modify the loss function of existing registration approaches and compare the results. To develop a suitable loss function, we come up with a novel way of posing registration as a multi-class classification problem.
%We test this hypothesis by modifying the loss functions of state-of-the-art registration networks. In order to develop a suitable loss function, we come up with a novel way of posing registration as a multi-class classification problem. 
Wherein, each point in one point cloud is classified as corresponding to a point in the other point cloud. These modified networks predict correspondences, from which pose parameters are then calculated using Horn's method~\cite{Horn87}. We show that the performance of each network that we modified is substantially improved (see Fig.~\ref{fig:intro_result}). Notably, these networks give more accurate registration results when faced with large initial misalignment and are more robust to partial point cloud data as compared to the original networks.

%%% old 
% Even though recent learning-based methods such as DCP~\cite{DCP} and RPMNet~\cite{RPMNet} also calculate correspondences as an intermediate step in order to calculate pose parameters, the networks are not explicitly trained to learn them. We show that their networks can register more accurately when explicitly trained to learn the correspondence.. Other works such as MVR~\cite{MVR} and DGR~\cite{Choy_2020_CVPR} introduce a two-step process of first finding a subset of correspondences and then classify each pair of correspondence as an inlier or an outlier. 
% We show that if correspondence assignment is treated as a multi-class classification instead, this two step process can be merged in a single step.
%%% old 

Even though recent learning-based methods such as DCP~\cite{DCP} and RPMNet~\cite{RPMNet} also calculate correspondence as an intermediate step in order to calculate pose parameters, the networks are not explicitly trained to learn them. We show that their networks can register more accurately when explicitly trained to learn the correspondence. 
% Other works such as MVR~\cite{MVR} and DGR~\cite{Choy_2020_CVPR} introduce a two-step process of first finding a subset of correspondences and then classify each pair of correspondence as an inlier or an outlier. Which is different from ours since we are classifying each point in a point cloud as belonging to a one of the classes \emph{i.e.} one of the point in other point clouds. 
% In fact in Sec.~\ref{PCR_as_multiclass}, we show that if correspondence assignment is treated as a multi-class classification problem then the two step process suggested by DGR~\cite{Choy_2020_CVPR} can be merged in a single step.

The key contributions of our work are listed as follows--
\begin{itemize}[noitemsep,topsep=0pt,parsep=0pt,partopsep=0pt]
    \item We provide a fundamental reasoning of why explicitly predicting correspondence provides better accuracy and results in faster convergence and verify it through extensive experimentation. 
    \item We introduce a new way of formulating point cloud registration as a multi-class classification problem and develop a suitable loss using point correspondence.
    % \item We demonstrate how existing network architectures can be tweaked to predict the point correspondences and hence can be used to register partial point clouds. 
    % \item We provide empirical proofs of the effectiveness of training with a correspondence-based loss function over conventional loss functions by rigorous experimentation on state-of-the-art methods.
    % \item Proposed approach has a faster convergence rate.
\end{itemize}

\section{Related Work}
\subsection{Conventional registration methods}
 Iterative Closest Points (ICP)~\cite{ICP} is one of the most popular methods for point cloud registration. ICP iteratively computes nearest neighbor correspondences and updates transformation parameters by minimizing the least-squares error between the correspondences~\cite{Horn87}. Over the years, several variants of ICP have been developed~\cite{rusinkiewicz2001efficient}. An important area of research in this space relates to efficient ways of finding correspondences, for example point-to-plane correspondences~\cite{chen1992object}, probabilistic correspondences~ \cite{segal2009generalized,Seth15,maier2010iterative,Myronenko10}, and feature-based correspondences~\cite{rangarajan1999rigid,stamos2003automated}.  These methods are locally optimal and hence perform poorly in the case of large misalignment.

For large misalignment, stochastic  optimization techniques have been developed such  as  genetic  algorithms~\cite{silva2005precision},  particle swarm  optimization~\cite{wachowiak2004approach},  particle  filtering~\cite{sandhu2009point,srivatsan2016multiple} etc.
Another category of methods that deal with large misalignment include globally optimal techniques. A popular approach is the globally optimal ICP (Go-ICP)~\cite{Yang13} that uses a branch and bound algorithm to find the pose. Recently, mixed integer programming has been used to optimize a cost function over transformation parameters and correspondences simultaneously ~\cite{srivatsan2019globally,izatt2020globally}. These methods have theoretical guarantees to reach global optima. The fact that they explicitly optimize over correspondences, motivates our work. 
%  \textcolor{red}{Write about 4pcs, 3D normal distribution transform, Fast Point Feature Histogram, color based registration, RANSAC.}

% To align point clouds at higher misalignment, various handcrafted feature descriptors such as FPFH ~\cite{rusu2009fast}, VFH\cite{muja2011rein}, GFPFH~\cite{rusu2009detecting}, OUR-CVFH~\cite{aldoma2012our}, etc.  are designed. These feature descriptors are then used to sample unique points and are aligned with methods such as RANSAC ~\cite{thomas2013stable} Or they are used to estimate possible correspondence and then pose parameters are calculated using closed form least square solutions like Horn's method ~\cite{Horn87}.

\subsection{Deep learning-based registration methods}
Some of the recent deep-learning based PCR methods train a network to directly predict the transformation between the input point clouds. PointNetLK~\cite{PointNetLK} aligns the point clouds by minimizing the difference between the PointNet~\cite{pointnet} feature descriptors of two input point clouds. PCRNet~\cite{PCRNet}, passes the concatenated global feature vectors through a set of fully connected layers to predict the pose parameters. These methods operate on global point cloud features and fail to capture the local geometrical intrinsics of the points. 

In order to capture local geometry, Deep Closest Point~\cite{DCP} learns to assign embedding to the points in each point cloud based on its nearest neighbours and attention mechanism. Further based on the similarity between the features, a correspondence matrix is generated which calculates transformation parameters that are used to define the loss function. DCP network architecture is iteratively used by PRNet~\cite{wang2019prnet} to align partial point clouds. This idea of using a correspondence predictor iteratively is also used by RPMnet~\cite{RPMNet}, where the network structure uses FGR~\cite{zhou2016fast} feature descriptor unlike DGCNN~\cite{wang2019dynamic} used by DCP and PRNet. 
% These methods can be together put as function $f$ with input as two sets of 3D points of $N_x$ and $N_y$ elements in no particular order and output transformation parameters such as quaternions or rotation vectors or rotation matrices along with translation vectors \emph{i.e.} $f: (\mathbb{R}^{3 \times N_x},\mathbb{R}^{3 \times N_y}) \xrightarrow[]{} SE(3)$. 
% 
% zhou2016fast

Some other methods such as deep global registration (DGR)~\cite{Choy_2020_CVPR} and multi view registraiton (MVR)~\cite{MVR}, follow a two step process -- (1) they find a set of correspondence pairs between two  sets of 3D points using fully convolution geometric features (FCGF)~\cite{choy2019fully}, and (2) these correspondence pairs are passed through a network which filters outliers. Note that DGR and MVR only find a subset of all possible correspondence pairs, and yet register more accurately than methods that directly predict pose parameters. This observation motivates us to study the effect of explicitly training a network to predict all point-point correspondences.
A key difference between the approach taken by MVR and DGR from our approach is that, they take point-pairs and classify them as inlier/outliers, while our approach classifies each point in one set (source) as belonging to one of multiple available classes (points in the other set) 

\section{Mathematical Formulation}

PCR is generally posed as an optimization problem. Consider two point clouds $\boldsymbol{X}  = [\boldsymbol{x}_1,\boldsymbol{x}_2,...,\boldsymbol{x}_{N_x}]\in \mathbb{R}^{3 \times N_x}$ and  $ \boldsymbol{Y}  = [\boldsymbol{y}_1,\boldsymbol{y}_2,...,\boldsymbol{y}_{N_y}] \in \mathbb{R}^{3 \times N_y}$ containining $N_x$ and $N_y$ points respectively, where $\boldsymbol{x}_i \in \mathbb{R}^{3}$ and $\boldsymbol{y}_i \in \mathbb{R}^{3}$ are the points in the respective point clouds and generally, $N_x \neq N_y$. The ground truth transformation, $\boldsymbol{R}^* \in SO(3)$ and $\boldsymbol{t}^* \in \mathbb{R}^{3}$, that aligns the two point clouds can be represented as 
\begin{equation}
    \boldsymbol{R}^*, \boldsymbol{t}^* = 
    \underset{\boldsymbol{R}, \boldsymbol{t}}{argmin}
    \left(\sum_{i=1}^{N_x}||\boldsymbol{R}\boldsymbol{x}_{i} + \boldsymbol{t} - \boldsymbol{y}_{\pi(\boldsymbol{x}_i)}||_2 \right),
\end{equation}
 where $||\dots||_2$ is the L2 norm and, $\pi$ denotes a function such that $\pi({\boldsymbol{x}_i})$ is the index of the point corresponding to $\boldsymbol{x}_i$ in $\boldsymbol{Y}$. This function can be represented by a binary matrix known as correspondence matrix $\boldsymbol{C}\in \mathbb{R}^{N_y \times N_x}$, where ${C_{i,j}}\in \{0,1\}$ $\forall$ $i\in [1,\dots,N_x]$ and $j\in [1,\dots,N_y]$. $\boldsymbol{C}_{j,i} = 1$ implies that $\boldsymbol{y}_j$ corresponds to $\boldsymbol{x}_i$. Or, $\boldsymbol{y}_j = \boldsymbol{y}_{\pi(\boldsymbol{x}_i)} = \boldsymbol{YC}_{:,i}$. Here $\boldsymbol{C}_{:,i}$ represents the $i^{th}$ column of  $C$. Since the correspondence is unknown, registration is restated as, 
\begin{equation}
    \boldsymbol{R}^*, \boldsymbol{t}^*, \boldsymbol{C}^* = 
    \underset{\boldsymbol{R}, \boldsymbol{t}, \boldsymbol{C}}{argmin}
    \left(\sum_{i=1}^{N_x}||\boldsymbol{R}\boldsymbol{x}_{i} + \boldsymbol{t} - \boldsymbol{YC}_{:,i}||_2 \right) \label{optimization}
\end{equation}

Where $\boldsymbol{C}^*$ is the ground truth correspondence matrix. Note that $\hat{\boldsymbol{Y}}=\boldsymbol{YC}$ denotes the rearranged $\boldsymbol{Y}$ such that $i^{th}$ point of $\boldsymbol{X}$ corresponds to $i^{th}$ point of $\hat{\boldsymbol{Y}}$.

\subsection{Robustness of correspondences Vs robustness of transformations}
\begin{figure}[t]
    \begin{center}
    \centering
        \includegraphics[width=0.7\linewidth]{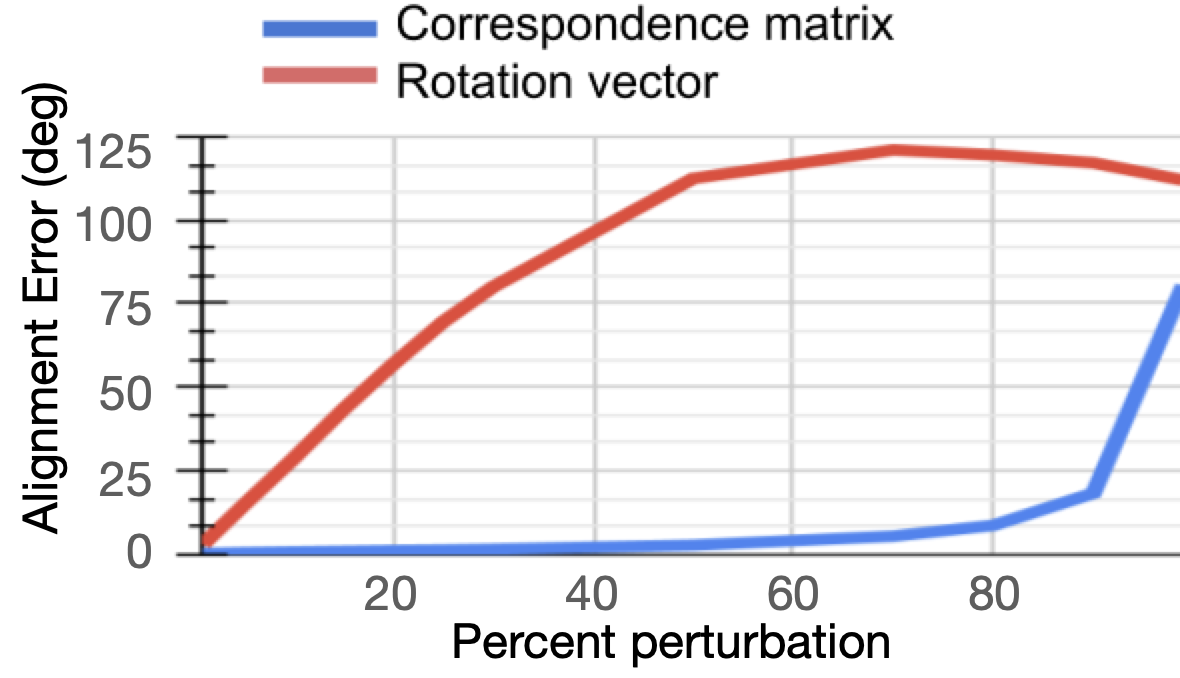}
    \end{center}
    \caption{Graph showing percentage of perturbation to `point-correspondence' and `rotation vector' vs alignment error. The plot shows that the alignment error is low even with as high as $40\%$ wrong correspondences. On the other hand the alignment error quickly increases with perturbation to the rotation parameters. Thus, we hypothesize that training the networks to learn correspondences would have better registration accuracy than learning pose parameters.}
    % \label{fig:long}
    \label{fig:percent_corruption}
\end{figure}
\label{robustness_correspondence}
To understand the effect of wrong correspondences on alignment, we perform the following experiment. We initially sample $n$ points randomly from a unit 3D cube and denote it as point cloud $\boldsymbol{X}$. We then transform $\boldsymbol{X}$ with a random but known rotation $\boldsymbol{R}^*$ to create point cloud $\boldsymbol{X'}$ = $\boldsymbol{R}^*\boldsymbol{X}$. For convenience, we convert $\boldsymbol{R}^*$ to a rotation vector form $\boldsymbol{v}^*\in\mathbb{R}^3$ and add $p\%$ corruption to generate $\boldsymbol{v}^{pert} = \boldsymbol{v}^{*} +  \boldsymbol{v}^{corrupt}$. We then calculate the error between $\boldsymbol{R}^{pert}$ and $\boldsymbol{R}^{*}$. This is noted as alignment error between corrupted rotation and ground truth rotation. We gradually increase the percentage corruption and calculate the corresponding alignment error (Figure \ref{fig:percent_corruption}). To observe the robustness of the correspondences, in another independent experiment, we randomly corrupt $p\%$ of the ground truth correspondence and calculate the resulting rotation matrix based on perturbed correspondences using Horn's method \footnote{Note that Horn's method is just one of many closed form approaches to obtain transformation given corresponding pairs of point clouds. The results will be identical if Horn's method is replaced by weighted SVD, or Arun's method~\cite{Arun87}.}~\cite{Horn87}.  The error between the ground truth rotation and perturbed rotation is then calculated. We gradually increase the percentage corruption and observe it's effect on the rotation error. We observe that even if $40 \%$ of the correspondences are wrong, the alignment error is $\approx 5^\circ$. 
%Thus correspondence parameters  is a more robust parameter for PCR as compared to transformations, and motivates us to leverage this robustness.

Based on this observation we hypothesize that if a network is trained explicitly to predict correspondence, the network will align point clouds more accurately than a network with similar architecture but trained to predict pose.
\section{PCR as multi-class classification}
\label{PCR_as_multiclass}
To test our hypothesis, we first develop a suitable loss function that can explicitly learn the correspondences. An obvious choice could be a mean square error or absolute error between predicted and ground truth correspondence but these loss functions do not provide any strong physical intuition about the correspondence.

We introduce a novel way of treating the task of correspondence assignment as a multi-class classification problem. We treat each point in $\boldsymbol{Y}$ to be a different class and each point in $\boldsymbol{X}$ belongs to one of the classes i.e. $N_x$ examples and $N_y$ classes. Note that each example needs to belong to at least one class but there can be classes with no corresponding example. This framework is particularly suitable to register partial point clouds where, $\forall \boldsymbol{x}_i,  \exists \boldsymbol{y}_j$ but converse need not be true. Note that this is fundamentally different from MVR~\cite{MVR}, where each correspondence pair is classified as a binary: inlier or outlier and the correspondence matrix constraints are not respected.  

We consider a general framework that first generates per point features $\boldsymbol{F}_{\boldsymbol{X}} = [\boldsymbol{f}_{x_1}, \boldsymbol{f}_{x_2}, ..., \boldsymbol{f}_{x_{N_x}}] \in \mathbb{R}^{N_e \times N_x }$ and $\boldsymbol{F}_{\boldsymbol{Y}} = [\boldsymbol{f}_{y_1}, \boldsymbol{f}_{y_2}, ..., \boldsymbol{f}_{y_{N_y}}] \in \mathbb{R}^{N_e \times N_y }$ for input point clouds $\boldsymbol{X}$ and $\boldsymbol{Y}$ where $\boldsymbol{f}_j \in \mathbb{R}^{N_e \times 1}$ and $N_e$ is the embedding space dimension. We generate a soft correspondence matrix based on a differentiable distance metric in the feature space~\footnote{The soft correspondence is similar to the matrix used in conventional registration approaches~\cite{Myronenko10, izatt2020globally, srivatsan2019globally,Seth15}, where every element of the correspondence matrix denotes the probability of matching.}.  The metric can be distance-based as introduced by MVR~\cite{MVR} or projection-based as suggested in DCP~\cite{DCP}. Without any loss of generality, we choose DCP's approach to generate a soft correspondence matrix \emph{i.e.} a matrix where each element denotes probability of a correspondence between point pairs as   $\boldsymbol{C} = softmax(\boldsymbol{F}_{\boldsymbol{Y}}^T\boldsymbol{F}_{\boldsymbol{X}})$. We compare $\boldsymbol{C}$ with a ground truth correspondence matrix $\boldsymbol{C}^*$.
We define the ground-truth correspondence as nearest neighbor of a point $\boldsymbol{x}_i$ in $\boldsymbol{Y}$ when $\boldsymbol{X}$ and $\boldsymbol{Y}$ are aligned. It is worth noting that this is not a reversible mapping \emph{i.e.} if $\boldsymbol{y}_1 \in \boldsymbol{Y}$ is the nearest neighbor of $\boldsymbol{x}_1$, it is possible that the nearest neighbor of $\boldsymbol{y}_1$ in $\boldsymbol{X}$ is $\boldsymbol{x}_j$, where $j\neq 1$. This adds a constraint on the correspondence matrix that each the sum of the elements of each column should add up to one.

The multi-class classification framework allows us to use a  cross-entropy loss, $L_{CE}$. This loss function implicitly applies the constraint that sum of the elements of a column should be one unlike binary cross entropy (BCE). For the sake of convenience of notation,  we define $\boldsymbol{C'}= \boldsymbol{F}_{\boldsymbol{Y}}^T\boldsymbol{F}_{\boldsymbol{X}}$
\begin{equation*}
    L_{CE} (\boldsymbol{C'}, \boldsymbol{C^*})= 
        -\sum_{i=1}^{N_x} log\left( 
    \frac{exp\left(\sum\limits_{j=1}^{N_y}\boldsymbol{C'}_{j,i}\boldsymbol{C^*}_{j,i}\right)}
        {\sum\limits_{j=1}^{N_y}exp(\boldsymbol{C'}_{j,i})} 
    \right) \label{cross_entropy_corr}
\end{equation*}

\begin{figure}[t]
    \begin{center}
        \includegraphics[width=0.8\linewidth]{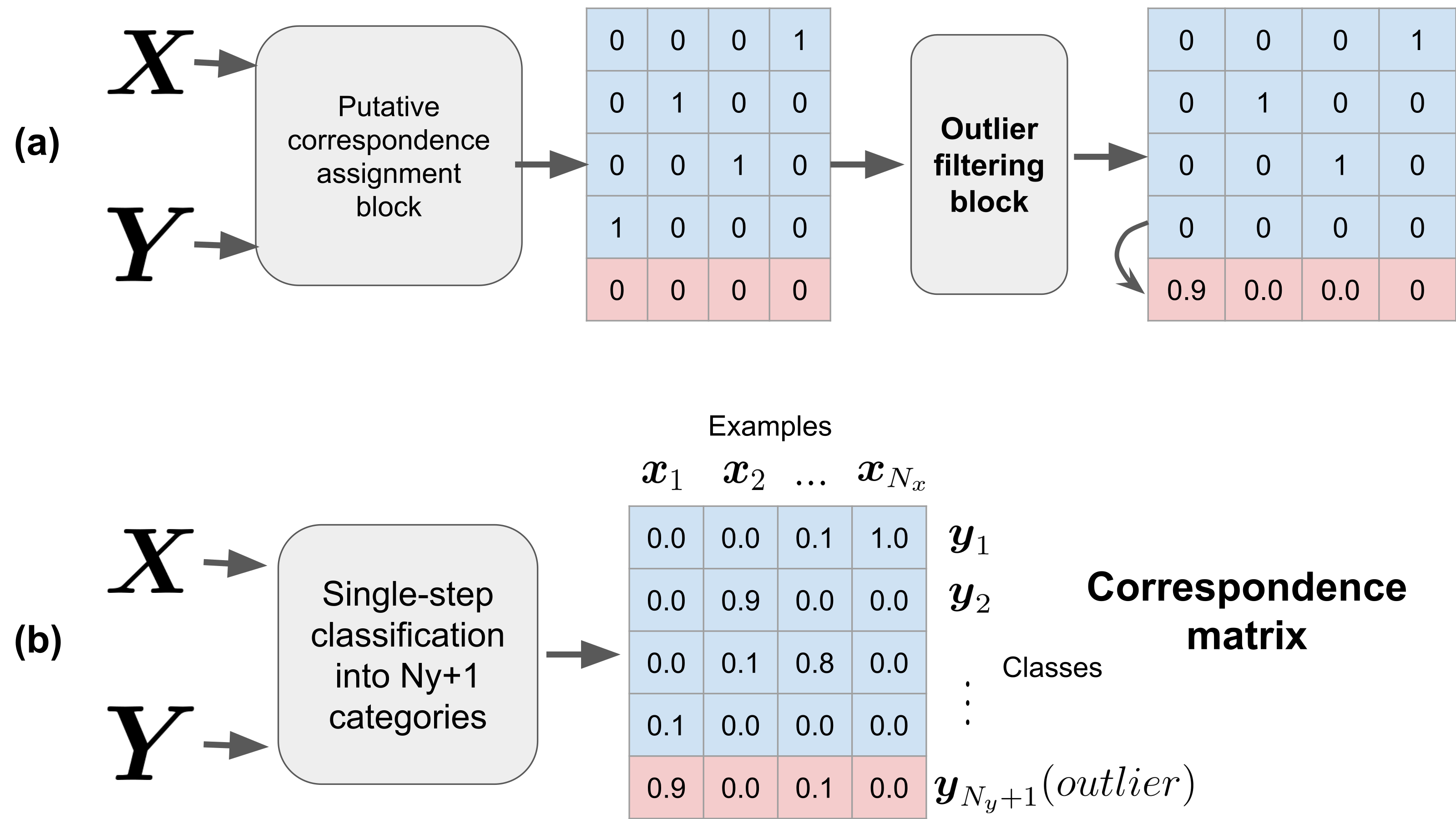}
    \end{center}
    \caption{a) Deep Global Registration~\cite{Choy_2020_CVPR}, b) Possible extension of our approach where two stage process of DGR can be merged into a single step thus resulting in a faster registration. Note that in our current work, networks output $\boldsymbol{C} \in \mathbb{R}^{N_y \times N_x}$ we assume that the data can be partial but there are no outliers}
    % \label{fig:long}
    \label{fig:DGR_vs_ours}
\end{figure}

While beyond the scope of this paper, it is worth noting that our framework can be further modified to add an additional class to classify an $\boldsymbol{x}_i$ as an outlier. If we incorporate an extra class for outliers in the correspondence matrix $\boldsymbol{C} \in \mathbb{R}^{N_y+1 \times N_x}$, this becomes a single-step generalised version of the two step process used by DGR (see Fig.~\ref{fig:DGR_vs_ours}).  

\section{Experimental setup}
\label{sec:experimental_setup}

% \textcolor{red}{In this section, we empirically want to analyze the effect of explicitly learning the correspondence Vs implicitly learning correspondence on registration accuracy. }
We consider RPMNet ~\cite{RPMNet}, DCP~\cite{DCP}, and PCRNet~\cite{PCRNet} to study the effects of training the network to learn correspondence vs training the network to learn pose parameters. Note that these methods were originally developed to register point clouds with small ($\mathit{\pm 45^\circ}$) initial misalignment. From here on we follow the notation that \emph{method} is the network trained with loss function suggested in the original paper while \emph{method\_corr} is trained using our loss function (cross-entropy on correspondence matrix). We train and test all these \emph{method}s and \emph{method\_corr}s on ModelNet40~\cite{wu20153d} dataset.

\begin{figure}[t]
    \begin{center}
        \includegraphics[width=0.9\linewidth]{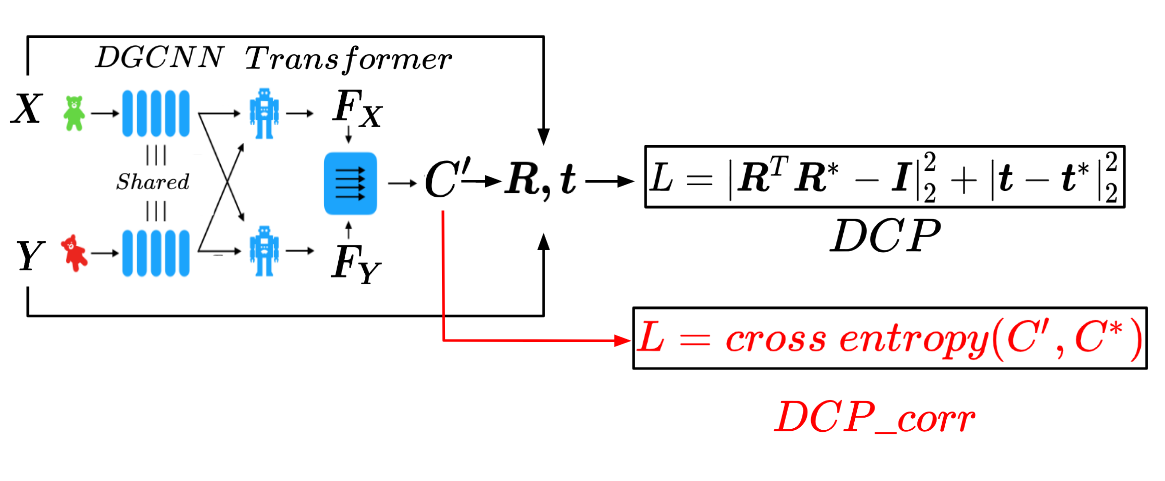}
    \end{center}
    \begin{center}
        \includegraphics[width=0.9\linewidth]{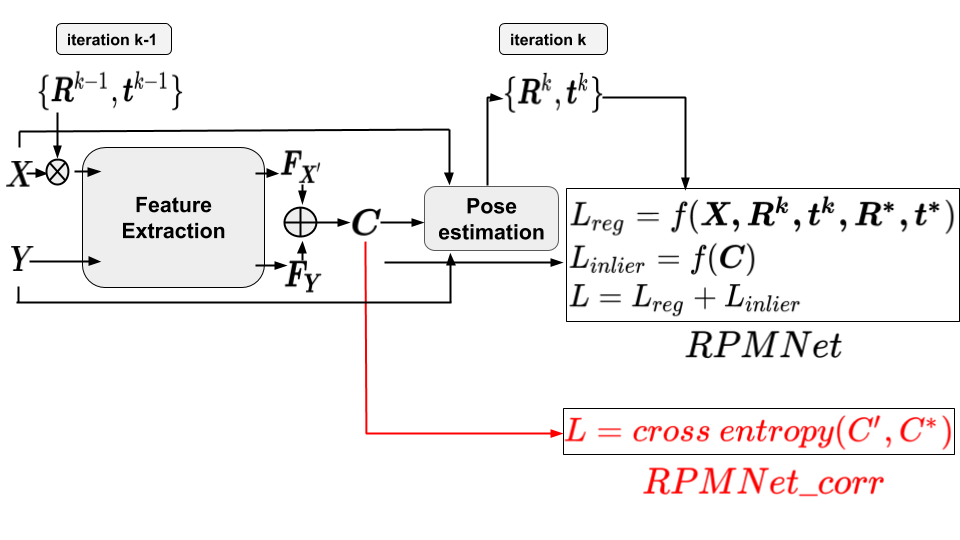}
    \end{center}
    
    \begin{center}
    \includegraphics[width=0.9\linewidth]{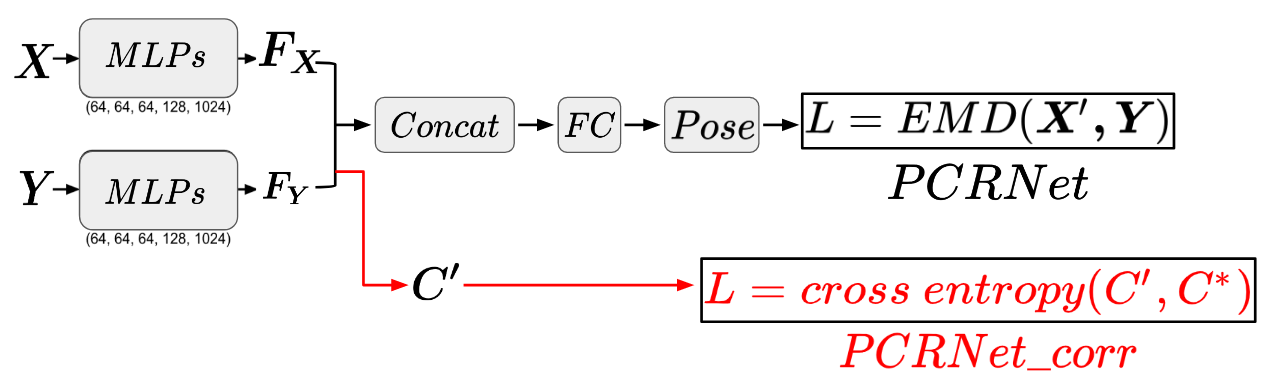}
    \end{center}
    \caption{DCP and RPMNet  architctures internally calculate the correspondence matrix $\boldsymbol{C}$ . This correspondence matrix is further used along with $\boldsymbol{X,Y}$ to calculate $\boldsymbol{R,t}$. In order to make these networks explicitly learn correspondence, we use $\boldsymbol{C}$ along with ground truth $\boldsymbol{C^*}$ to calculate cross entropy loss. Since PCRNet does not explicitly calculate $\boldsymbol{C}$, we modify the network architecture and compare the PointNet's per-point features to generate the correspondence matrix.} 
    \label{fig:DCP_arch}
\end{figure}

 DCP~\cite{DCP} and RPMNet ~\cite{RPMNet} generate an implicit correspondence based on the similarity between per-point features of the input point clouds (Fig. ~\ref{fig:DCP_arch}). This intermediate correspondence is used to find the transformation parameters between input point clouds using Horn's method~\cite{Horn87} and weighted SVD method ~\cite{RPMNet} respectively. For a network to implicitly learn correspondence, we define the loss as a function of the output transformation as suggested by the respective $method$. While to explicitly learn the correspondence, we define the loss as a function of intermediate correspondence as defined in~ Sec. \ref{cross_entropy_corr}.

% We use default splits provided by ModelNet40 dataset for training and testing. The dataset contains 3D CAD models of 40 different object categories.  Each CAD model consists of 2048 points and the objects are fitted into a unit cube.  

We sample $n$ number of points from a point cloud chosen from training data and denote this as point cloud $\boldsymbol{X}$. We generate a copy of $\boldsymbol{X}$ and shuffle the order of points to generate $\boldsymbol{Y'}$. To sample a rotation, we randomly choose a unit vector in $\mathbb{R}^3$ and an angle $\theta \in \mathcal{U}(-\theta_0, \theta_0)$, this axis and angle is used to generate a rotation vector which is further transformed into a ground-truth rotation matrix $\boldsymbol{R^*}$. Here, $\theta_0$ depends upon the specific experiment and $\mathcal{U}(a,b)$ denotes a uniform distribution in the range $[a,b]$. Further we generate a ground-truth translation vector $\boldsymbol{t^*} \in [\mathcal{U}(-0.5,0.5), \mathcal{U}(-0.5,0.5), \mathcal{U}(-0.5,0.5)]$. Now $\boldsymbol{Y'}$ is transformed with $\boldsymbol{R^*}$ and $\boldsymbol{t^*}$ to generate $\boldsymbol{Y}$.
To generate the ground-truth correspondence matrix $\boldsymbol{C^*}$, we find the nearest neighbour of each $\boldsymbol{x}_i \in \boldsymbol{X}$ in $\boldsymbol{Y'}$. If $\boldsymbol{y'}_j \in \boldsymbol{Y}$ is the nearest neighbour of $\boldsymbol{x}_i$ then $\boldsymbol{C}^*(j,i)$ is set to $1$ and other elements of $i^{\text{th}}$ column are set to $0$. 

To generate the partial point clouds, we randomly choose a plane passing through the centroid of the original point cloud of source ($\boldsymbol{X}$). We then randomly choose either up or down directtion of the plane and remove a predetermined number of points from the source farthest from the plane. 

\subsection{DCP Vs DCP\_corr}

DCP uses DGCNN~\cite{wang2019dynamic} features along with transformer network-based attention and co-attention mechanism to generate interrelated per point features of a point cloud. These features are used to generate probability distribution of source points on the target points matrix $\boldsymbol{C}$. They further calculate an intermediate representation of target point cloud $\boldsymbol{Y}$ as $\boldsymbol{\hat{Y}} = \boldsymbol{CY}$. DCP uses Horn's method to estimate the rotation matrix $\boldsymbol{R}$ and translation   $\boldsymbol{t}$ which minimizes the distance between corresponding points of $\boldsymbol{\hat{Y}}$ and $\boldsymbol{X}$. The loss for DCP is defined as 
\begin{equation}
    L_{DCP} = ||\boldsymbol{R}^T\boldsymbol{R}^* - \boldsymbol{I}||_2^2 +  ||\boldsymbol{t} - \boldsymbol{t}^*||_2^2    \label{Eq:DCP_loss}  
\end{equation}
For all the comparisons between DCP and DCP\_corr, we use \mbox{learning rate = 0.001} as recommended by DCP. 

DCP\_corr uses the correspondence matrix obtained in the intermediate step and compares it with ground truth correspondence using cross entropy
\begin{equation}
    L_{DCP\_corr} = cross~entropy(\boldsymbol{C},\boldsymbol{C^*})    \label{DCP_corr_loss}  
\end{equation}

\subsection{RPMNet vs RPMNet\_corr}

% {\bf RPMNet:}
% RPMNet is a recent network designed for robust PCR. As against DCP, RPMNet can handle noisy points and partial visibility. 
% The RPMNet framework consists of feature extraction network, parameter prediction network, estimatiion of correspondence matrix and further computing the transformation parameters using the estimated correspondences.
RPMNet follows an iterative procedure. In each iteration, the point clouds $\boldsymbol{X}$ and $\boldsymbol{Y}$ and transformation from previous iterations are passed into the feature extraction network which computes point-wise features.
% and absolute position of the points as well as the point normals. The outlier and the annealing parameters are computed from the two point  clouds using the parameter prediction network. 
The extracted features are then used to estimate the correspondence matrix which is further refined using Sinkhorn~\cite{sinkhorn1964relationship} normalization layer in an unsupervised manner.
In order to estimate the transformation parameters, the target points $\boldsymbol{Y}$ are weighted with the correspondence matrix weights $\boldsymbol{C}$ to obtain putative source correspondences $\hat{\boldsymbol{Y}}=\boldsymbol{YC}$. RPMNet\_corr uses this correspondence matrix to define the cross entropy loss (see Fig. \ref{fig:DCP_arch}). RPMNet evaluates transformation parameters $\boldsymbol{R,t}$ based on $\boldsymbol{X,\hat{Y}}$ and $\boldsymbol{C}$. These transformation parameters are then used to define the primary loss function $L_{reg}$, 
% Each correspondence pair ($\boldsymbol{x}_j,\hat{\boldsymbol{y}}_j$) is weighted with $w_j = \sum_{k=1}^{K} C_{jk}$. These weighted correspondence pairs are used to compute the rotation $\boldsymbol{R}$ and translation $\boldsymbol{t}$ using differentiable SVD.

\begin{equation}
      L_{reg} = \frac{1}{N_x} \sum_{i=1}^{N_x} |(\boldsymbol{R}^*\boldsymbol{x}_i + \boldsymbol{t}^*) - (\boldsymbol{R}\boldsymbol{x}_i + \boldsymbol{t}) |_1
\end{equation}

% In order to increase the number of inliers, an additional inlier loss function ($L_{inlier}$) is given as,

% \begin{equation}
%     L_{inlier} = - \frac{1}{N_x} \sum_{i=1}^{N_x} \sum_{j=1}^{N_y} \boldsymbol{C}_{ij} - \frac{1}{N_y} \sum_{j=1}^{N_y} \sum_{i=1}^{N_x}  \boldsymbol{C}_{ij}
% \end{equation}

RPMNet uses an additional unsupervised loss function $L_{inlier}$ which forces the network to predict majority of the correspondences as inliers. These two loss functions together form $L_{RPMNet} = L_{reg} + L_{inlier}$.

% Though the inlier loss function uses the correspondence matrix, it only supervises the number of points as inliers. However, $L_{RPMNet}$ lacks in its ability to completely supervise the correspondence matrix.

% Hence, we train RPMNet with our proposed classification loss function (equation 3). We call this approach RPMNet\_corr. 

Both RPMNet and RPMNet\_corr are trained with the same hyper-parameters (as recommended in~\cite{RPMNet}), except for the learning rate. RPMNet\_corr is trained with an initial learning rate of 0.01 which decays upto 0.0001 during training. We tried a higher learning for both the methods but training of RPMNet is unstable for higher learning rates. 
% Hence, we train RPMNet with a steady learning rate of 0.0001 as recommended by the authors.

% Specific implementation details about RPMNet(Rahul). (Learning rate)
% \begin{itemize}
%     \item Clean 45 RPMNet and RPMNet\_modified. R\_rmse, \%corr of both.
%     \item Clean 180 RPMNet and RPMNet\_modified. R\_rmse \%corr of both.
%     \item clean Partial 45 RPMNet and RPMNet\_modified. R\_rmse \%corr of both.
% \end{itemize}

% Specific implementation details about DCP(Tejas).

\subsection{PCRNet Vs PCRNet\_Corr}
PCRNet is a correspondence-free network that estimates registration parameters given a pair of input point clouds ($\boldsymbol{X}$ and $\boldsymbol{Y}$). As shown in Fig.~\ref{fig:DCP_arch}, PCRNet uses PointNet~\cite{pointnet} as a backbone to compute the point-wise features of each input point cloud arranged in a siamese architecture. 
In order to avoid input permutations, a symmetry function (max-pool) is operated on point-wise features to obtain a global feature vector ($\in \mathbb{R}^{1x1024}$). PCRNet concatenates the global feature vectors of both the inputs and uses a set of fully connected layers to regress the registration parameters. Rather than defining the loss function on the ground truth transformation, PCRNet uses chamfer distance (CD) as the loss function,
\begin{align}
    CD(\boldsymbol{X}, \boldsymbol{Y}) = & \frac{1}{N_x} \sum_{\boldsymbol{x}_i \in \boldsymbol{X}} \min_{\boldsymbol{y}_j \in \boldsymbol{Y}} \Vert \boldsymbol{x}_i - \boldsymbol{y}_j\Vert_2 + \nonumber \\ & \frac{1}{N_y} \sum_{\boldsymbol{y}_j \in \boldsymbol{Y}} \min_{\boldsymbol{x}_i \in \boldsymbol{X}} \Vert \boldsymbol{y}_j - \boldsymbol{x}_i\Vert_2
\end{align}
CD calculates the average closest distance between the template $\boldsymbol{X}$ and the point cloud obtained by applying predicted transformation on $\boldsymbol{Y}$.

Even though PCRNet uses an unsupervised loss function, CD is a function of $\boldsymbol{X}$, $\boldsymbol{Y}$, $\boldsymbol{R}$ and $\boldsymbol{t}$. In other words, the training of PCRNet again depends on the accuracy of $\boldsymbol{R}$, $\boldsymbol{t}$ when compared to the ground truth. 
%Given the ground truth R,t, CD loss becomes exactly zero.

% In order to show that our loss is better, we modify the network architecture of PCRNet. After computing point-wise features from PointNet for X and Y, we estimate the correspondence matrix by projecting the features as X*Y. This not only allows the PCRNet to produce correspondences but also reduces the number of trainable parameters.

% In this experiment, we train original PCRNet and PCRNetcorr using the point clouds from ModelNet40 dataset. Both the networks are trained with two setting having an initial misalignment within a range of [-45, 45] and [-180, 180] for 100 points in each of the input point cloud. From Fir.~\ref{}, we clearly observe that PCRNetcorr shows less rotation and translation error as compared to PCRNet even with lesser learnable parameters.

% Specific implementation details about PCRNet(Vinit).
\begin{figure*}[t]
\begin{center}
% \fbox{\rule{0pt}{2in} \rule{.9\linewidth}{0pt}}
\includegraphics[width=0.8\textwidth, height=0.4\textheight]{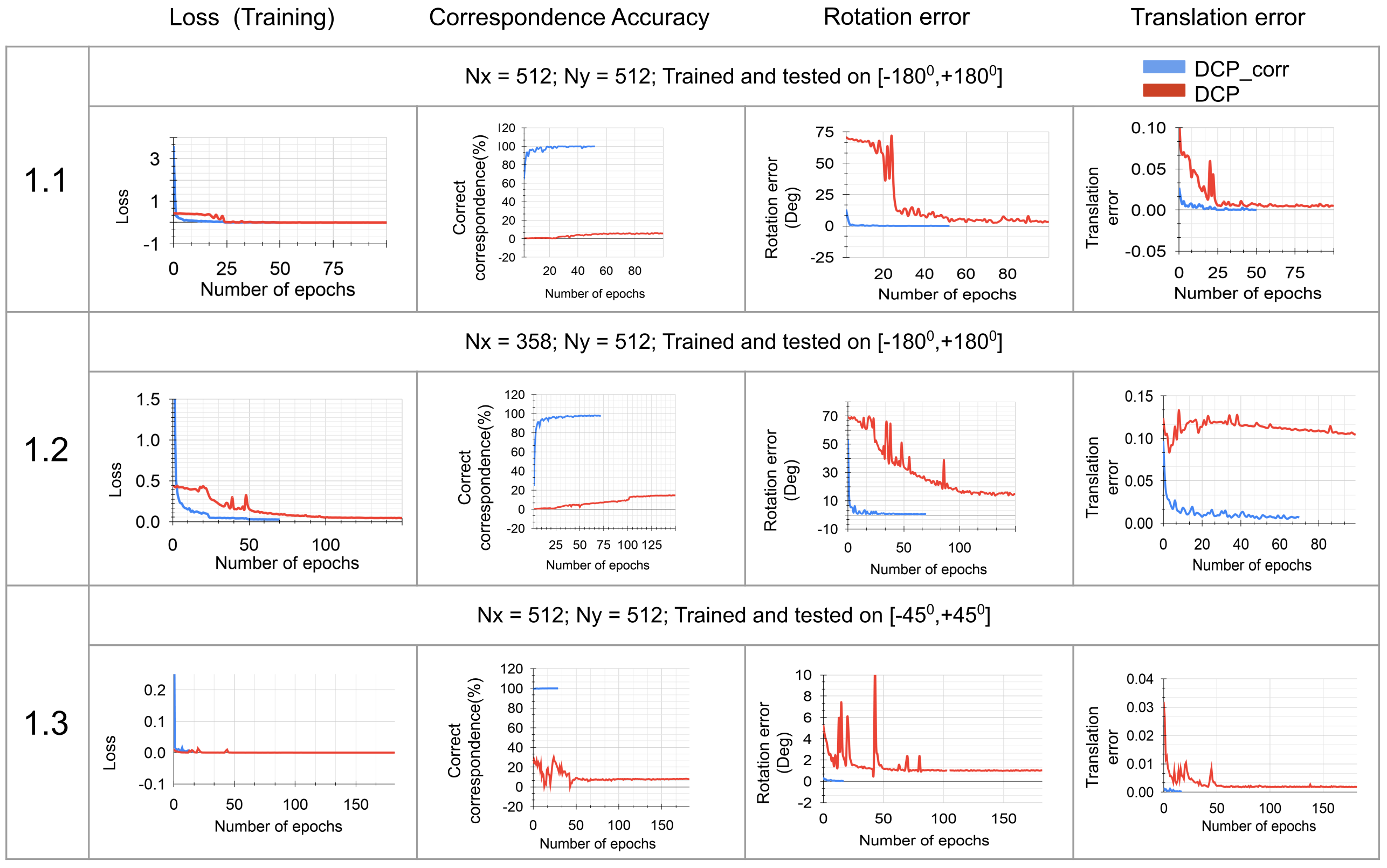}
\end{center}
   \caption{Results of experiments on DCP vs DCP\_corr. 
   %As a general trend in all these experiments, we observe that even though eventually both DCP(Red) and DCP\_corr(Blue) losses converge, DCP does not learn the correspondences accurately. For instance in Exp 1.1 after approx 100 epochs when DCP has converged, the correspondence accuracy is $\approx 5\%$ while that of DCP\_corr is $\approx 99 \%$. 
   }
\label{fig:dcp_plot_results}
\end{figure*}
\section{Results}
In this section, we present results of different existing approaches, referred to as $method$, and provide comparisons to versions of those approaches modified by training using our correspondence based loss -- referred to as $method\_corr$. We specifically highlight the improvement shown by $method\_corr$ compared to $method$ to large initial misalignment errors as well as ability to register partial point-clouds.

\setlength{\tabcolsep}{1pt}
\renewcommand{\arraystretch}{0.75}
\begin{table}[htbp]
\caption{Effect of initial misalignment on registration accuracy}
\centering
\begin{tabular}{ c |c| c |c| c } 
\toprule
Rotation & \multicolumn{2}{ |c|}{Rotation MAE (deg)} &\multicolumn{2}{|c }{Correspondence (\%)}\\
\cline{2-5}
range (deg)&DCP&DCP\_corr&DCP&DCP\_corr\\
\hline
 0-30 & 0.99 & 0.005 & 8.90 & 99.99\\ 
 30-60 & 1.55 & 0.008 & 6.12 & 99.97\\  
 60-90 & 1.69 & 0.010 & 5.78 & 99.96 \\  
 90-120 & 1.56 & 0.010 & 5.69 & 99.96 \\ 
 120-150 & 1.62 & 0.010 & 5.66 & 99.95 \\
 150-180 & 1.64 & 0.010 & 5.60 & 99.96\\
 \hline
\end{tabular}
\label{tb:DCP_rotation_range}
\end{table}
\subsection{ DCP Vs DCP\_corr}

% \begin{figure}[b]
%     \begin{center}
%         % \fbox{\rule{0pt}{2in} \rule{0.9\linewidth}{0pt}}
%         \includegraphics[width=\linewidth]{figures/DCP_table.png}
%     \end{center}
%     \caption{Observing effect of initial misalignment on registration accuracy between DCP and DCP\_corr }
%     % \label{fig:long}
%     \label{fig:DCP_rotation_range}
% \end{figure}

% \begin{figure}[b]
%     \begin{center}
%         % \fbox{\rule{0pt}{2in} \rule{0.9\linewidth}{0pt}}
%         \includegraphics[width=0.8\linewidth]{figures/DCP_pointclouds.png}
%     \end{center}
%     \caption{Visualizing registration accuracy of DCP\_corr Vs DCP on partial point clouds for arbitrary misalignment }
%     % \label{fig:long}
%     \label{fig:dcp_modified_yhat}
% \end{figure}
% \setlength{\tabcolsep}{2pt}
% \renewcommand{\arraystretch}{1}
% \begin{table}[htbp]
% \caption{Effect of initial misalignment on registration accuracy}
% \centering
% \begin{tabular}{ c |c| c |c| c } 
% \toprule
% Rotation & \multicolumn{2}{ |c|}{Rotation MAE (deg)} &\multicolumn{2}{|c }{Correspondence (\%)}\\
% \cline{2-5}
% range (deg)&DCP&DCP\_corr&DCP&DCP\_corr\\
% \hline
%  0-30 & 0.99 & 0.005 & 8.90 & 99.99\\ 
%  30-60 & 1.55 & 0.008 & 6.12 & 99.97\\  
%  60-90 & 1.69 & 0.010 & 5.78 & 99.96 \\  
%  90-120 & 1.56 & 0.010 & 5.69 & 99.96 \\ 
%  120-150 & 1.62 & 0.010 & 5.66 & 99.95 \\
%  150-180 & 1.64 & 0.010 & 5.60 & 99.96\\
%  \hline
% \end{tabular}
% \label{tb:dcp_results}
% \end{table}

The authors of DCP, consider 1024 points in all of their experiments. Due to limited GPU space, we re-ran all the DCP experiments using 512 points with the same hyper-parameters including learning rate for both. We sample rotations from $SO(3)$ with rotation vectors instead of Euler angles. This helped us to train DCP even for large misalignment. 
For different experimental settings Fig. \ref{fig:dcp_plot_results} shows the comparison between DCP and DCP\_corr. The first column shows that every training procedure converged. Second show the accuracy of correspondence estimation of both the methods. Third column shows rotation error as an RMSE over Euler angle error and fourth column denotes translation error.

\textbf{Experiment 1.1} We have $N_x = 512$ points in the source and $N_y = 512$ points in the target. The initial misalignment between them is uniformly sampled from $SO(3)$ while the translation is bound in cube of unit size centered on origin. As observed in Fig.~\ref{fig:dcp_plot_results} we can see that DCP\_corr converges faster than DCP and is more accurate. 

\textbf{Experiment 1.2} The results of this section are visualized in Fig.~\ref{fig:intro_result}. We have $N_x = 358$ points in the source and $N_y = 512$ points in the target. The source point cloud is made partial as described in Sec.~\ref{sec:experimental_setup} . We observe that even though DCP's loss function converges, the RMSE rotation error is $14.7 ^\circ$ while the rotation error of DCP\_corr is $0.51 ^\circ $. This can be considered as an empirical evidence that multi-class classification approach can deal with partial data without any major modification to the network architecture.

\textbf{Experiment 1.3} In this experiment, we compare DCP to DCP\_corr for the specific task DCP was developed for, \emph{i.e.} full-to-full point cloud registration for initial misalignment in the range of $[-45^\circ,+45^\circ]$. We observe that both the networks converge, and the rotation accuracy of DCP and  DCP\_corr are $1.036^\circ$ and $0.034^\circ$ respectively.

\textbf{Experiment 1.4} In this experiment we observe the effect of initial misalignment on registration accuracy of DCP and DCP\_corr trained for arbitrary initial misalignment (Table~\ref{tb:DCP_rotation_range}). For this experiment, we set the translation to zero and only allow a rotational misalignment between the input point clouds. We observe that DCP\_corr always registers more accurately than DCP, which is attributed to the remarkably high percentage of correct correspondence.   

 \begin{figure*}[htbp]
\begin{center}
% \fbox{\rule{0pt}{4in} \rule{.9\linewidth}{0pt}}
\includegraphics[width=0.8\textwidth]{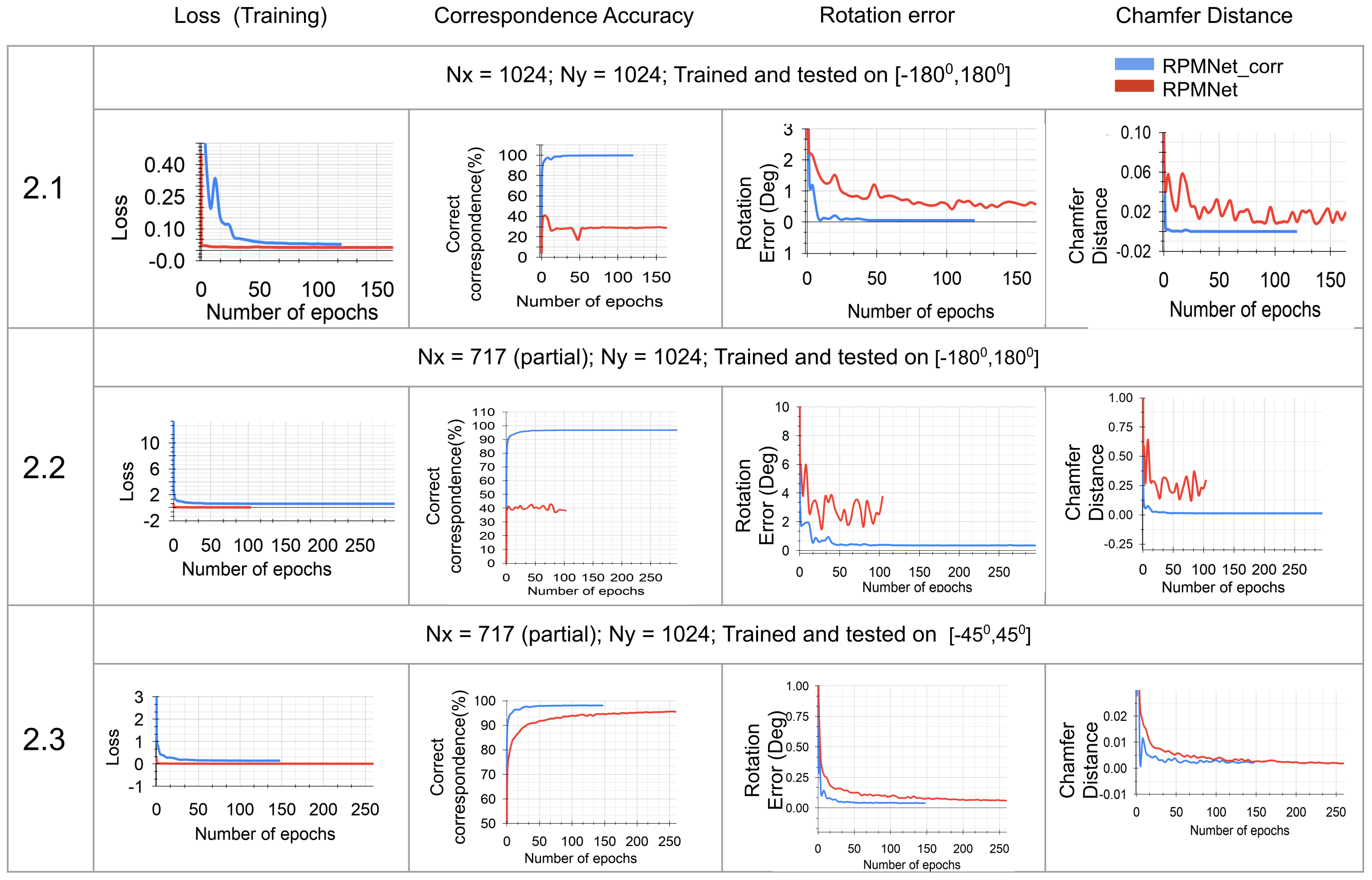}
\end{center}
   \caption{Results of experiments on RPMNet vs RPMNet\_corr }
\label{fig:RPMNet_chart}
\end{figure*}

\begin{figure*}[htbp]
    \begin{center}
        \includegraphics[width=0.8\textwidth]{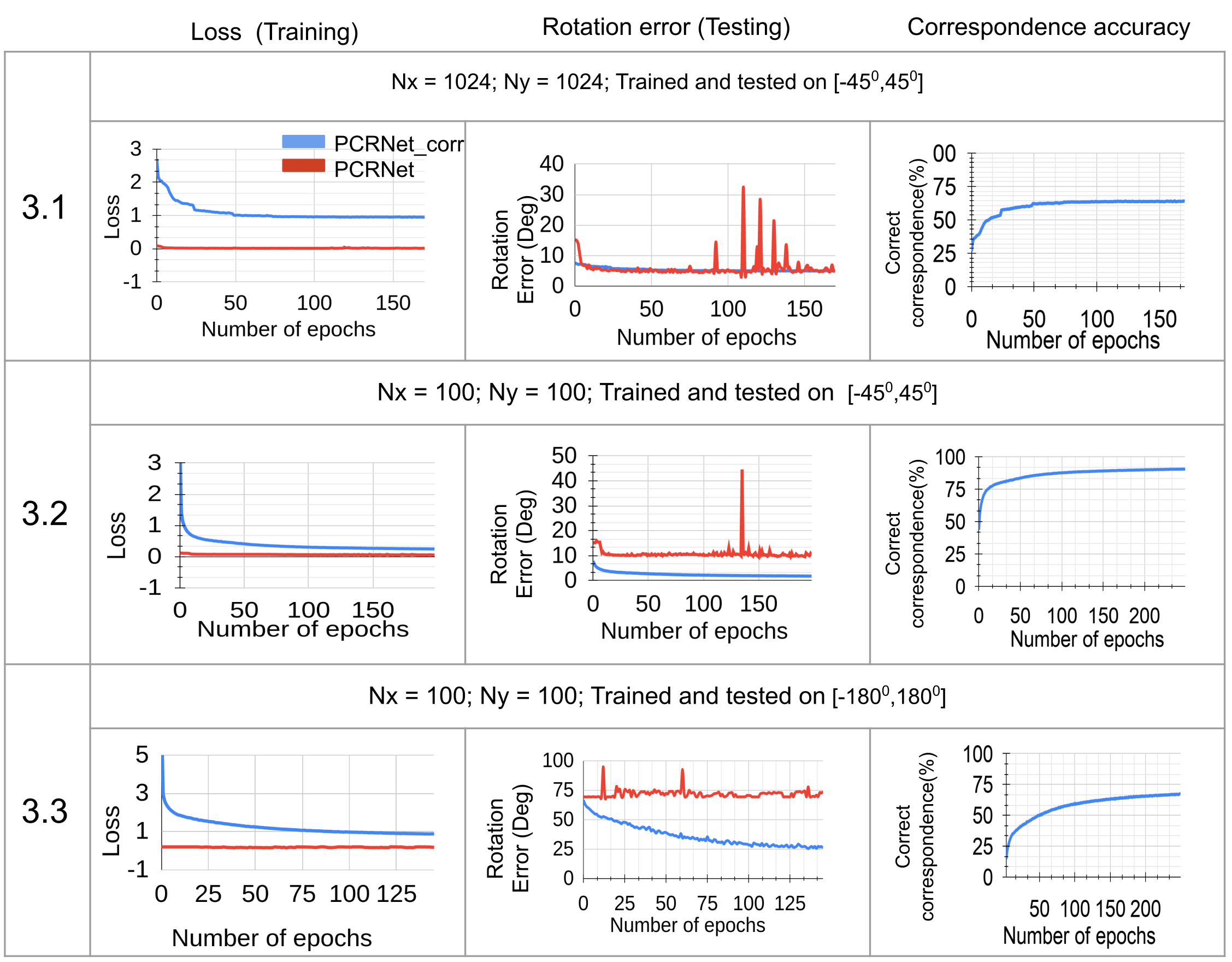}
    \end{center}
    \caption{ Results of experiments on PCRNet Vs PCRNet\_corr}
    % \label{fig:long}
    \label{fig:PCRNet_chart}
\end{figure*}

\subsection{ RPMNet Vs RPMNet\_corr}

We present the comparisons between RPMNet and RPMNet\_corr in Fig. \ref{fig:RPMNet_chart}. Unlike the previous experiment with DCP, the rotation error metric used to evaluate these experiments is the mean absolute anisotropic rotation error also known as axis angle error. We chose this metric to be compliant with the choice of the authors of RPMNet~\cite{RPMNet}. Likewise, we present the Chamfer distance (CD) between registered point clouds, in the fourth column of  Fig. \ref{fig:RPMNet_chart}, as suggested by the authors of RPMNet~\cite{RPMNet}.
The initial misalignment in translation is sampled uniformly between $[-0.5,0.5]$. 
% Specifically, we conduct experiments for the following four cases:

\textbf{Experiment 2.1}
In this experiment, both the point clouds have $N_x = N_y = 1024$ points.  The misalignment between these clouds is uniformly sampled from $SO(3)$. 
% As seen in Fig. 8 (row 1), the \% of correct correspondence is converged to a  much higher value in case of RPMNet\_corr (99.7\%) as compared to RPMNet trained with the original loss function (28.9\%). Since the model can estimate the correspondence matrix correctly, 
It can be observed that the rotation error converges faster and to a lower value of $0.059^\circ$ with RPMNet\_corr as compared to an error of $0.56^\circ$ for RPMNet.

\textbf{Experiment 2.2}
% Real world scans of the objects from the sensors are partial in nature due to self-occlusion. This experiment tests the ability of our proposed approach to register a partial source point cloud to a full target cloud. 
To test the ability of multi-class classification approach to handle partial point clouds, in this experiment we generate the partial source point cloud by retaining 70\% of the points above a random plane such that $N_x = 717$ and $N_y = 1024$. 
We carry out this experiment with uniform sampling from $SO(3)$. 
% 2.1. Fig. 8 (row 2) shows the difference in \% of correct correspondence predicted, 38.5\% against 96.9\% for RPMNet and RPMNet\_corr respectively. 
Note that, even though one of the key features of RPMNet is the ability to deal with partial point clouds, RPMNet\_corr has higher registration accuracy of $0.34^\circ$ compared to  $3.79^\circ$ of RPMNet. 
% $3.79^\circ$ rotation error, the same network trained using our proposed loss function achieves more than 10 times lesser error ($0.34^\circ$).

\textbf{Experiment 2.3}
RPMNet is specifically designed  for $[-45^\circ,+45^\circ]$ initial misalignment. Even in this range, we observe that RPMNet\_corr converges faster and registers more accurately. We also observe that eventually RPMNet reaches $96\%$ correspondence accuracy. We believe that the RPMNet's Sinkhorn algorithm along with the unsupervised loss on correspondence ($L_{inlier}$), pushes the intermediate correspondence matrix towards the ground truth correspondence matrix in an unsupervised manner.  

\textbf{Experiment 2.4} In this experiment we study the effect of initial misalignment on the registration accuracy of RPMNet and RPMNet\_corr. Both the networks are trained with arbitrary initial misalignment in the range of $[-180^\circ, +180^\circ]$ between the input point clouds. In this experiment, we set the translation to zero and  only allow a rotational misalignment between the input point clouds. 
We calculate Mean Absolute Error (MAE) between  predicted and ground truth rotation in Euler angles.  
We observe from Table~\ref{tb:RPMNet_rotation_range} that the MAE for rotation is always lower for RPMNet\_corr when compared to RPMNet. 
%We also observe that RPMNet\_corr performs better even with CD metric.

% {\bf Experiment 2.4}
% Finally, we also test our approach against partial to full point cloud registration with $[-45^\circ,+45^\circ]$ initial misalignment. The partial point clouds are sampled in the same way as done in experiment 2.1. RPMNet predicts 95.4\% correct correspondences with $0.061^\circ$ rotation error while RPMNet\_corr predicts 98.1\% of correspondences correct with $0.039^\circ$ rotation error.

\setlength{\tabcolsep}{3pt}
 \renewcommand{\arraystretch}{1.5}

\begin{table}[htbp]
\centering
\caption{Effect of initial misalignment on registration accuracy }
 \resizebox{0.65\textwidth}{!}{\begin{minipage}{\textwidth}
\begin{tabular}{ c |c| c |c| c |c|c} 
\toprule
 & \multicolumn{2}{ |c|}{Rotation } &\multicolumn{2}{|c }{Correspondence}&\multicolumn{2}{|c }{Chamfer distance }\\
Rotation & \multicolumn{2}{ |c|}{ MAE (deg)} &\multicolumn{2}{|c }{ (\%)}&\multicolumn{2}{|c }{MSE$\times$ 1E-5 }\\
\cline{2-7}
range (deg)&RPMNet&RPMNet\_corr&RPMNet&RPMNet\_corr&RPMNet&RPMNet\_corr\\
\hline
 0-30 & 0.52 & 0.011 & 26.96 & 98.28& 8.51 & 0.56\\  
 30-60 & 0.58 & 0.013 & 26.97 & 98.28& 8.48 & 0.56\\  
 60-90 & 0.69 & 0.015 & 26.96 & 98.28 & 8.51 & 0.55\\  
 90-120 & 0.89 & 0.26 & 26.96 & 98.18 & 8.49 & 0.68\\  
 120-150 & 1.01 & 0.47 & 26.97 & 98.15 & 9.36 &0.12\\
 150-180 & 0.96 & 0.66 & 26.97 & 98.07& 8.68 & 0.79\\
 \hline
\end{tabular}
\label{tb:RPMNet_rotation_range}
  \end{minipage}}
\end{table}

\subsection{ PCRNet Vs PCRNet\_corr}
The results showing the comparison between PCRNet and PCRNet\_corr are shown in \ref{fig:PCRNet_chart}. We only provide a rotational misalignment between the input point clouds.

\textbf{Experiment 3.1} We consider $N_x=N_y=1024$ points for both the input point clouds. We train PCRNet with the hyper-parameters recommended in~\cite{PCRNet} and compare it with PCRNet\_corr. Note that PCRNet\_corr has fewer tunable parameters than PCRNet due to the removal of MLPs. We observe that both the approaches converge to $ \approx 5^\circ$ rotation accuracy. Based on the results of this experiment, we believe that PCRNet lacks depth or number of parameters to achieve higher accuracy. Another reason to believe this is, even after doing a thorough hyper-parameter search, we could not achieve correspondence accuracy of $\geq 70\%$. 

\textbf{Experiment 3.2} In this experiment, we have two point clouds with 100 points each. The initial misalignment between them is in the range of $[-45^\circ, 45^\circ]$. We observe that PCRNet converges to a rotation accuracy of $9.97 ^\circ$ compared to $1.8 ^\circ$ of PCRNet\_corr. 
% It is not possible to compare percent correspondences learnt by PCRNet since it does not calculate them anywhere unlike RPMNet and DCP.

\textbf{Experiment 3.3} We repeat the previous experiment but use an initial misalignment that is uniformly sampled from $SO(3)$. We observe that PCRNet\_corr outperforms PCRNet and is able to learn correspondences and the rotation accuracy reaches $21^\circ$ at the end of 250 epochs.

\section{Conclusion and Future Work}
In this paper we demonstrate that higher registration accuracy can be achieved if a network is trained to explicitly learn correspondences instead of learning them implicitly by training on registration parameters. This paper adds to the ever-increasing body of work demonstrating how carefully selecting the desired output of a data-driven approach can lead to drastic improvements in performance. We observe faster convergence, higher registration accuracy and ability to register partial point clouds when networks are explicitly trained to learn correspondence instead of pose parameters. We also developed a new way to approach registration as a multi-class classification task.

While in this work we have limited ourselves to results on ModelNet40,  we plan to extend it to real world datasets such as 3DMatch~\cite{zeng20173dmatch} and Sun3d~\cite{xiao2013sun3d}. In addition, future work will involve extended the multi-class classification approach to deal with outliers in the point clouds.

\section*{Supplementary}
\subsection*{A. Registration in the presence of outliers}
% As mentioned in Sec. 4, we can extend our multi-class classification approach to register in the presence of outliers. 

To extend multi-class classification approach to filter outliers, we consider a general framework that first generates per point features $\boldsymbol{F}_{\boldsymbol{X}} = [\boldsymbol{f}_{x_1}, \boldsymbol{f}_{x_2}, ..., \boldsymbol{f}_{x_{N_x}}] \in \mathbb{R}^{N_e \times N_x }$ and $\boldsymbol{F}_{\boldsymbol{Y}} = [\boldsymbol{f}_{y_1}, \boldsymbol{f}_{y_2}, ..., \boldsymbol{f}_{y_{N_y}}] \in \mathbb{R}^{N_e \times N_y }$ for input point clouds $\boldsymbol{X}$ and $\boldsymbol{Y}$. Here $\boldsymbol{f}_j \in \mathbb{R}^{N_e \times 1}$ and $N_e$ is the embedding space dimension. In the absence of outliers, we predicted the correspondence matrix (probability of each source point to belong to one of the target points)  as 

$$\boldsymbol{C} = softmax(\boldsymbol{F}_{\boldsymbol{Y}}^T\boldsymbol{F}_{\boldsymbol{X}}) \in \mathbb{R}^{N_y \times N_x}$$ 

In presence of outliers, we want to classify a source point to belong to one of the target points or as an outlier. \emph{i.e.} we need  $N_y + 1$ number of classes. To get an extra outlier class for classification we use a feature vector embedding $\boldsymbol{f_{Y_O}} \in \mathbb{R}^{N_e \times 1}$ that can suitably represent an outlier. With such embedding for outliers, we can predict an outlier variant of correspondence matrix $\boldsymbol{C}^O \in \mathbb{R}^{N_y + 1 \times N_x}$ as 

$$\boldsymbol{C}^O = softmax([\boldsymbol{f}_{y_1}, \boldsymbol{f}_{y_2}, ..., \boldsymbol{f}_{y_{N_y}}, \boldsymbol{f_{Y_O}}]^T
\boldsymbol{F}_{\boldsymbol{X}})$$    

Note that for $i^{\text{th}}$ source point, $\boldsymbol{C}^O_{1,i}, \boldsymbol{C}^O_{2,i}, ... ,  \boldsymbol{C}^O_{N_y,i}$ denotes the probability of the point belonging to $\boldsymbol{y}_1,\boldsymbol{y}_2,  ... , \boldsymbol{y}_{N_y}$ respectively and $\boldsymbol{C}^O_{N_y + 1,i}$ denotes the probability of it being an outlier. 

In case that a point is predicted as an outlier, we want the probability of it being classified as one of the source points, as less as possible. In order to do so, we want the outlier embedding to have large-negative projections on all the target point embeddings so that after the softmax operation probabilities of outlier to be classified as an inlier will be close to zero. There can be various ways to obtain such embedding. For preliminary experiments, we generate such embedding $\boldsymbol{f_{Y_O}}$ as,

\begin{equation}
    \boldsymbol{f_{Y_O}} = 
    \underset{\boldsymbol{f}}{argmin} 
    \left( ||\boldsymbol{F_{Y}}^T \boldsymbol{f} - b  \mathbf{1}_{N_y}||_2 \right) \nonumber
\end{equation}

Here $\mathbf{1}_{N_y}$ denotes a column vector of ones of length $N_y$ and $b$ is a scalar. We empirically choose $b$ to be $-1$.

\section*{B. Experiment - DCP Vs DCP\_corr in the presence of outliers }
We take $N_x = 512$ and $N_y = 512$ with initial misalignment in the range of $[-45^\circ,+45^\circ]$ for both DCP and DCP\_corr. Then position of $10\%$ points from $\boldsymbol{X}$ is randomly corrupted to be a  random point in a unit cube centered at the origin. To generate the outlier variant of ground-truth correspondence matrix $\boldsymbol{C^{O*}}$, we find the nearest neighbour of each $\boldsymbol{x}_i \in \boldsymbol{X}$ in $\boldsymbol{Y'}$ and it's distance to the nearest neighbor. If $d_i$ is less than a predefined threshold then $\boldsymbol{y'}_j \in \boldsymbol{Y}$ is denoted as nearest neighbour of $\boldsymbol{x}_i$ by setting $\boldsymbol{C}^{O*}(j,i)$ to $1$ and other elements of $i^{\text{th}}$ column to be $0$. If $d_i$ is greater than the predefined threshold, then $i^\text{th}$ source point is marked as an outlier by setting $\boldsymbol{C}^{O*}(N_y+1,i)$ to $1$ and other elements of $i^{\text{th}}$ column to be $0$.

Further DCP\_corr is trained to minimize the loss function $cross~entropy(\boldsymbol{C^O},\boldsymbol{C^{O*}})$ as mentioned in Sec. 4. The last row of  $\boldsymbol{C^O}$ is used as outlier weights and a weighted SVD operation is performed to obtain transformation parameters (refer [6] for details). For DCP\_corr, we observe an  rotation error (RMSE) of $1.485^\circ$ and an  translation error (RMSE) of $0.000138$ after 15 epochs. The outlier filtering by DCP\_corr can be visualized in (Fig. 8).  On the other hand, DCP is trained with the mean squared error loss on transformation parameters. We observe a rotation error of $6.704^\circ$ and translation error of 0.519 units. The effect of outliers on DCP vs DCP\_corr can be observed in the Fig. 9. 

\begin{figure}[t]
    \begin{center}
        \includegraphics[width=0.9\linewidth]{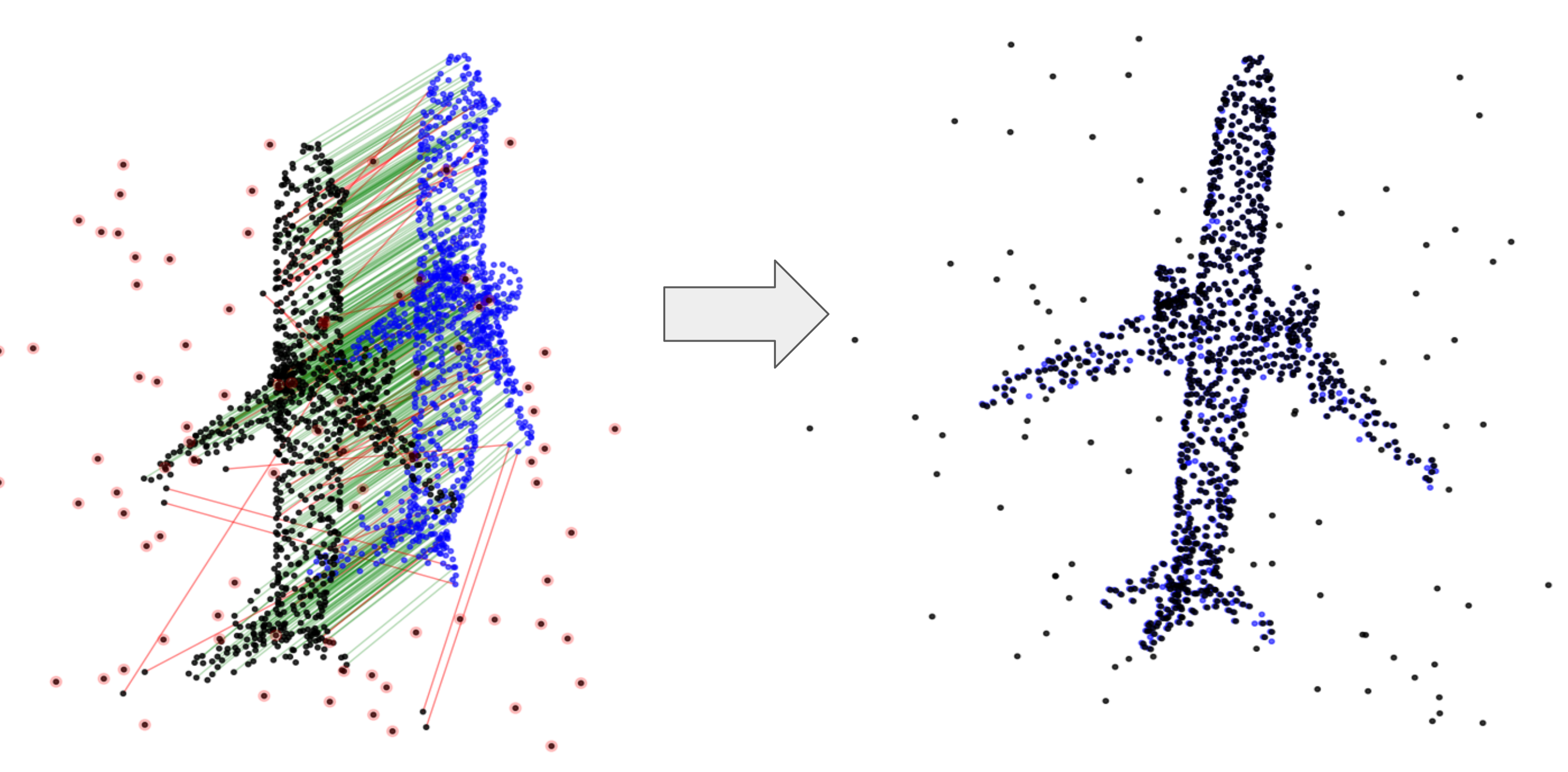}
    \end{center} 
    \begin{center}
        \includegraphics[width=0.9\linewidth]{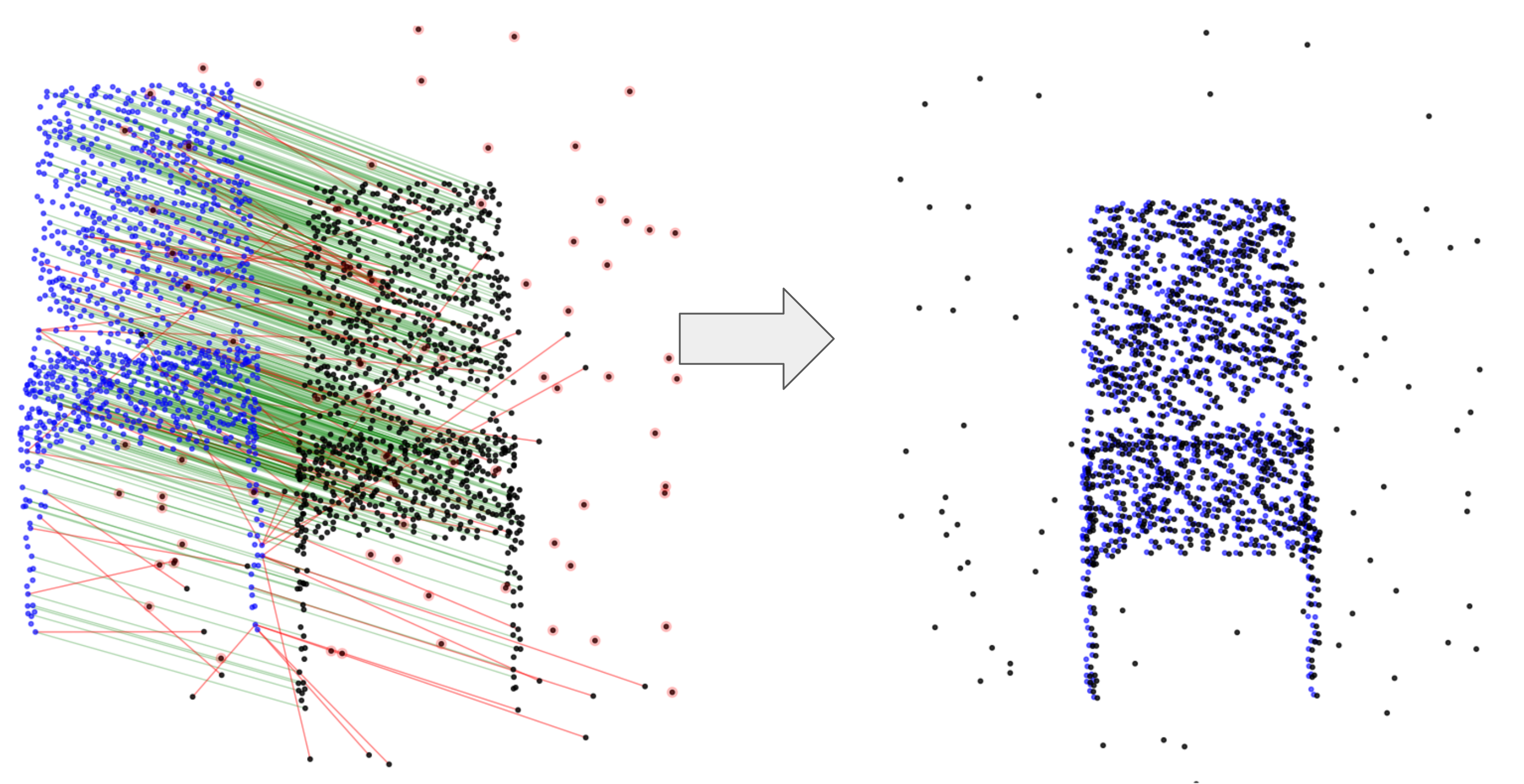}
    \end{center} 
    Figure. 8. Left: Initially misaligned point clouds. Black points are source, blue points are target. Red lines denote correspondence wrongly predicted by DCP\_corr while green lines denote correct predictions. Black points circled by red spheres denote points marked as outliers by the network. Right: Registered point clouds using predicted correspondence matrix. 
    
    \label{fig:intro_result}
\end{figure}

\begin{figure*}[t]
    \begin{center}
        \includegraphics[width=\textwidth]{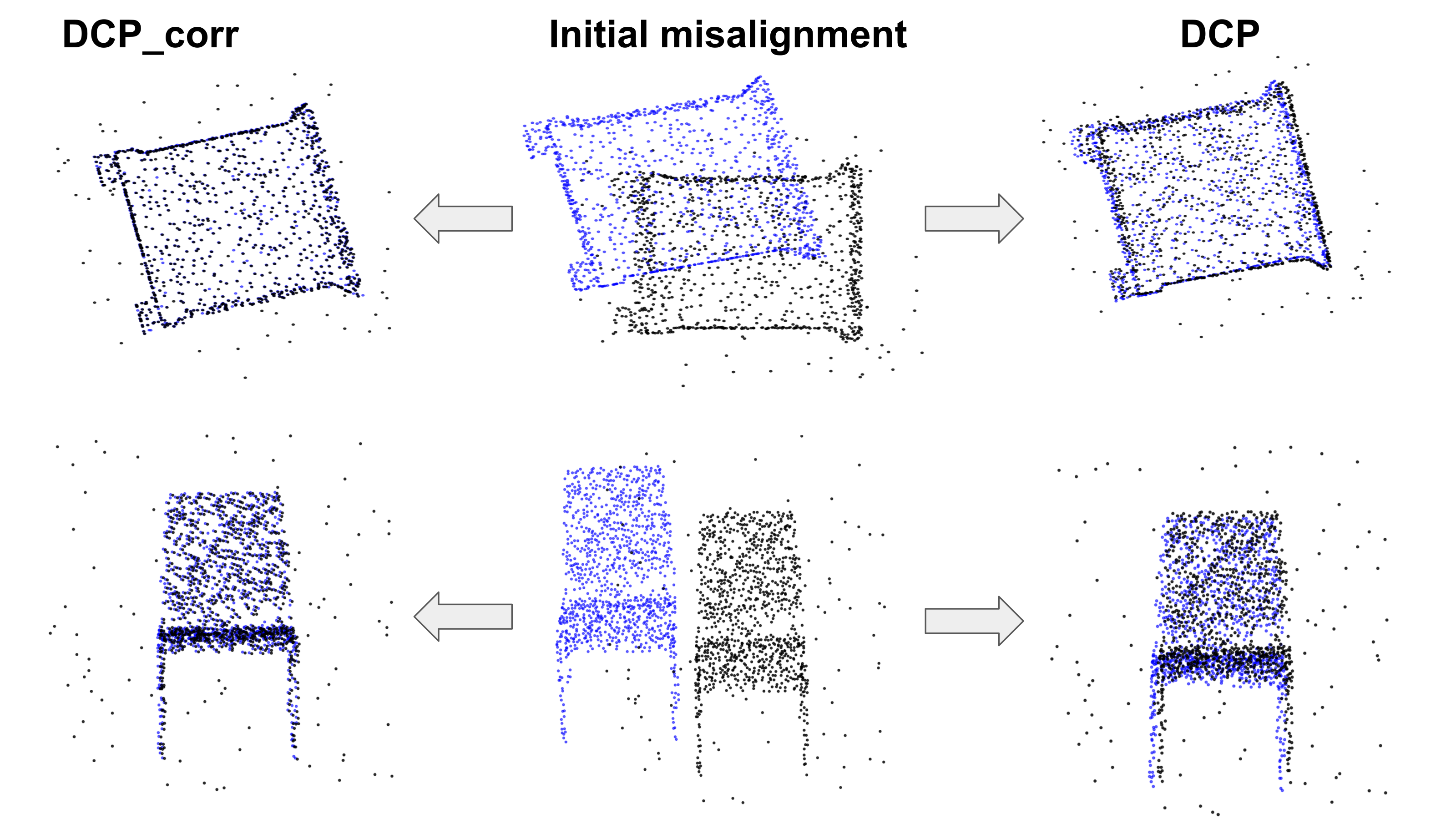}
    \end{center}
    \begin{center}
        \includegraphics[width=\textwidth]{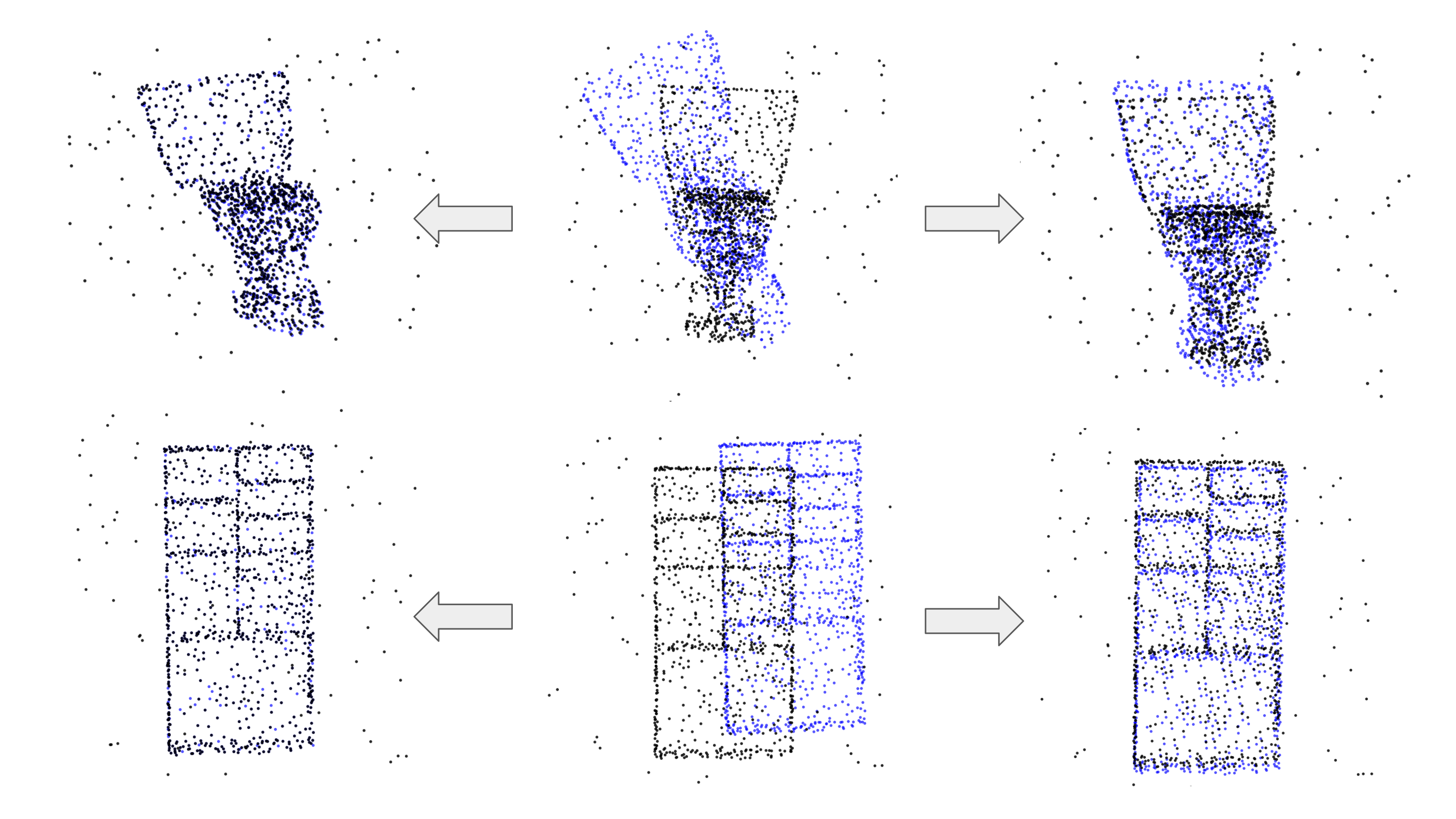}
        Figure 9. Visualization of DCP\_corr and DCP registration in presence of outliers
    \end{center}
 
\end{figure*}

{\small
\bibliographystyle{ieee}
\bibliography{egbib}
}

\end{document}